\documentclass[letterpaper, 10 pt, conference]{ieeeconf}  

\IEEEoverridecommandlockouts                              

\usepackage{graphicx}
\usepackage{amsmath,amssymb,bm}
\usepackage{mathtools}
\usepackage{subcaption}
\usepackage{hyperref}
\usepackage{cleveref}
\usepackage{siunitx}
\usepackage{float}
\usepackage{array}
\usepackage{tabularx}
\begin{document}

\title{\LARGE \bf
AFD-CAMLs: Agile Force-Distribution-Aware Planning and Control for Cable-Suspended Aerial Multi-Lifting Systems
}

    \author{Antreas Kourris, Sihao Sun$^*$
\thanks{The authors are with the Faculty of Mechanical Engineering, Delft University of Technology.}
\thanks{\textsuperscript{*}Corresponding author:
        {\tt\small s.sun-2@tudelft.nl}}}

\maketitle

\newcommand{\antreas}[1]{{\color{green}[Antreas: #1]}}
\newcommand{\sihao}[1]{{\color{red}[Sihao: #1]}}

\begin{abstract}
Multiple UAVs can cooperatively transport heavy payloads while controlling
their position and orientation. Trajectory-based methods offer high agility
while satisfying system constraints, but can produce uneven force distributions
when the tension-to-wrench allocation is redundant or ill-conditioned,
particularly under geometric mismatch and low-level tracking errors. We propose
a hybrid planning-and-control framework to address this problem. A global
planner generates payload trajectories and cable-force references by exploring
the allocation null space under a prescribed internal-force setting. These
references augment the cost of a centralized local planner, promoting feasible
force distributions while generating trajectories for all UAVs. An admittance
filter then compares the planned forces with onboard cable-tension estimates
and adjusts the kinematic references to improve force tracking in degenerate or
near-degenerate configurations. Simulations and experiments involving four to
ten UAVs demonstrate more balanced tension distributions during both hovering
and demanding agile maneuvers, without compromising agility or payload-tracking
performance.
\end{abstract}


\section{INTRODUCTION}
Teams of unmanned aerial vehicles (UAVs) can cooperatively transport payloads
beyond the capacity of a single vehicle
\cite{Michael2011CooperativeManipulation,Sreenath2013CooperativePayload}.
Adding UAVs increases lifting capacity but can make the cable-force distribution
non-unique. Accurate payload-pose tracking therefore does not ensure that the
load is shared safely and evenly among the vehicles. This distinction is
especially important for cable-suspended systems: the net force and moment on
the payload determine its motion, whereas internal-force components can change
the individual cable tensions without changing that net wrench. Excessive
tension can saturate or destabilize a vehicle, while insufficient tension can
cause cable slackness and invalidate the assumed system dynamics.

Existing methods can broadly be divided into position- and force-based
approaches. Quasi-static position-based methods determine UAV configurations
from payload poses, cable geometry, and static equilibrium
\cite{Michael2011CooperativeManipulation,Manubens2013MotionPlanning}. These
methods are effective for slow motion but neglect dynamic coupling and do not
guarantee feasible cable tensions. Force-based methods compute the required
payload wrench from a dynamic model and allocate it among the cables
\cite{Lee2018Geometric,Sanalitro2020FullPose,li2023nonlinear}, allowing
secondary objectives such as tension or internal-force regulation
\cite{Geng2022LoadDistribution}. Most retain a cascaded wrench-allocation and
inner-loop structure whose time-scale separation limits agile operation. Their
force commands also depend on nominal payload and cable parameters. The
resulting distribution is therefore sensitive to uncertainty in payload mass
and inertia, attachment locations, and cable lengths
\cite{Sanalitro2020FullPose,Sun2025AgileCable}.

\begin{figure}[t]
    \centering
    \includegraphics[width=\linewidth]{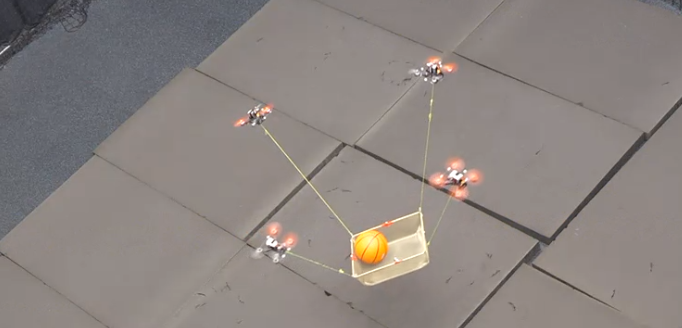}
    \caption{Snapshot of four UAVs manipulating a cable-suspended payload containing an
    additional unknown load using the proposed AFD-CAMLs framework.}
    \label{fig:eye-catcher}
\end{figure}

Recent work introduced a \emph{trajectory-based approach} that replaces the
conventional cascaded force controller with a centralized whole-body nonlinear
model predictive control (NMPC) planner and onboard trajectory-tracking
controllers \cite{Sun2023FullPoseNMPC,Sun2025AgileCable}. The planner predicts the coupled
load-cable dynamics and generates dynamically feasible trajectories for all
UAVs while enforcing thrust, cable-tautness, collision-avoidance, and
obstacle-avoidance constraints. Agility is improved because the onboard
controllers track complete kinematic references rather than realizing
instantaneous force commands through separate control layers. This formulation
also allows the planner to anticipate coupled payload and cable motion over a
finite horizon instead of allocating a wrench independently at each instant.

For four or more UAVs, however, multiple cable-force distributions can produce
the same payload wrench. The planner generates kinematic trajectories from a
nominal geometric model, and the low-level controllers do not directly enforce
the corresponding tensions. Attachment-point and cable-length errors, combined
with trajectory-tracking errors, can therefore shift the physical system toward
a different force equilibrium. The payload may still track accurately while
some UAVs become overloaded and other cables approach slackness. Improving
payload-pose tracking alone cannot correct this internal-force component because
it does not affect the resultant payload wrench. The problem becomes most
pronounced when the tension-to-wrench allocation matrix is degenerate or nearly
degenerate, because small geometric or tracking errors can then produce large
changes in the individual tensions. A practical solution must therefore select
a feasible distribution during planning and regulate its realization without
sacrificing the agility of the trajectory-based architecture.

This work extends the trajectory-based framework~\cite{Sun2025AgileCable} with
explicit planning and feedback regulation of the cable-force distribution. An
internal-force-aware global planner explores the null space of the payload
wrench-allocation matrix to generate feasible cable-force and direction
references. These references are incorporated into the centralized NMPC cost so
that the locally planned trajectories account for the desired force
distribution. Because the cable forces are still realized indirectly through
vehicle motion, an admittance filter compares each planned tension with an
onboard estimate and converts the error into position, velocity, and
acceleration corrections. The resulting hybrid motion--force architecture
combines predictive whole-body planning with local force feedback.

The three main contributions are:
\begin{enumerate}
    \item an internal-force-aware global planner that preserves the required
    payload wrench while generating tension-feasible cable-force and direction
    references through null-space optimization;
    \item a force-reference-aware NMPC formulation and an admittance filter that
    uses onboard tension estimates to regulate the realized force distribution
    through local kinematic corrections; and
    \item simulation studies with four to ten UAVs and real-world experiments with four UAVs (Figure~\ref{fig:eye-catcher}) that
    demonstrate improved force distribution under geometric and inertial-model
    mismatch while preserving agile payload tracking.
\end{enumerate}
\section{Preliminaries}

\subsection{Notation}

Bold lowercase letters denote vectors, bold uppercase letters denote matrices,
and non-bold symbols denote scalars. The inertial frame is denoted by
\(\mathcal{I}\), the payload-fixed frame by \(\mathcal{L}\), and the body-fixed
frame of UAV \(i\) by \(\mathcal{B}_i\). Unless indicated otherwise, vectors
are expressed in \(\mathcal{I}\). The rotation matrices
\(\mathbf{R},\mathbf{R}_i\in SO(3)\) map coordinates from
\(\mathcal{L}\) and \(\mathcal{B}_i\), respectively, to \(\mathcal{I}\).
For \(\bm{a}\in\mathbb{R}^3\), \([\bm{a}]_\times\) is the skew-symmetric
matrix satisfying \([\bm{a}]_\times\bm{b}=\bm{a}\times\bm{b}\).
The Moore--Penrose pseudoinverse and Euclidean norm are denoted by
\((\cdot)^\dagger\) and \(\|\cdot\|\), respectively, and
\(\|\bm{a}\|_{\mathbf{Q}}^2=\bm{a}^{\mathsf{T}}\mathbf{Q}\bm{a}\).
The superscript \(g\) denotes a global-planner reference, whereas the subscripts
\(\mathrm{cmd}\), \(\mathrm{ref}\), and \(\mathrm{obs}\) denote commanded,
controller-reference, and observed quantities, respectively.

\subsection{Multi-lifting system model}

We consider \(n>3\) unidirectional-thrust UAVs transporting a rigid payload
through massless, inextensible cables, following the load--cable model in
\cite{Sun2023FullPoseNMPC,Sun2025AgileCable}. The payload has mass \(m_L\),
inertia matrix \(\mathbf{J}_L\), position \(\bm{p}\), velocity \(\bm{v}\),
orientation \(\mathbf{R}\), and body-frame angular velocity \(\bm{\omega}\).
The \(i\)th cable has length \(l_i\) and a payload attachment point
\(\bm{\rho}_i\), expressed in \(\mathcal{L}\). We define
\(\bm{s}_i\in\mathbb{S}^2\) as the direction from this attachment point to UAV \(i\).
Thus, when the cable is taut,

\begin{equation}
\bm{p}_i
=
\bm{p}+\mathbf{R}\bm{\rho}_i+l_i\bm{s}_i,
\qquad
\bm{f}_i=T_i\bm{s}_i,
\qquad T_i>0,
\label{eq:cable_kinematics}
\end{equation}
where \(\bm{p}_i\) is the UAV position and \(\bm{f}_i\) is the cable force
acting on the payload. The equal and opposite force \(-\bm{f}_i\) acts on the
UAV. Under this convention, the payload dynamics are

\begin{equation}
\begin{aligned}
\dot{\bm{p}}&=\bm{v},~~~~
m_L\dot{\bm{v}}=m_L\bm{g}+\sum_{i=1}^{n}\bm{f}_i,\\
\dot{\mathbf{R}}&=\mathbf{R}[\bm{\omega}]_\times,~~~~
\mathbf{J}_L\dot{\bm{\omega}}
+[\bm{\omega}]_\times\mathbf{J}_L\bm{\omega}
=\sum_{i=1}^{n}
[\bm{\rho}_i]_\times\mathbf{R}^{\mathsf{T}}\bm{f}_i,
\end{aligned}
\label{eq:payload_dynamics}
\end{equation}
where \(\bm{g}\) is the gravity vector. UAV \(i\), with mass \(m_i\), collective
thrust \(u_i\), and body thrust axis \(\bm{e}_3=[0,0,1]^\top\), satisfies

\begin{equation}
\dot{\bm{p}}_i=\bm{v}_i,
\qquad
m_i\dot{\bm{v}}_i
=
m_i\bm{g}+u_i\mathbf{R}_i\bm{e}_3-\bm{f}_i.
\label{eq:uav_translation}
\end{equation}

Stacking the cable forces as
\(\bm{f}=[\bm{f}_1^\top,\ldots,\bm{f}_n^\top]^\top\), the force and moment
required by the payload dynamics can be written as

\begin{equation}
\mathbf{P}(\mathbf{R})\bm{f}=\bm{w},
\qquad
\mathbf{P}(\mathbf{R})
=
\begin{bmatrix}
\mathbf{I}_3 & \cdots & \mathbf{I}_3\\
[\mathbf{R}\bm{\rho}_1]_\times & \cdots &
[\mathbf{R}\bm{\rho}_n]_\times
\end{bmatrix},
\label{eq:wrench_mapping_matrix}
\end{equation}
where

\begin{equation}
\bm{w}
=
\begin{bmatrix}
m_L(\dot{\bm{v}}-\bm{g})\\
\mathbf{R}\!\left(
\mathbf{J}_L\dot{\bm{\omega}}
+[\bm{\omega}]_\times\mathbf{J}_L\bm{\omega}
\right)
\end{bmatrix},
\label{eq:required_payload_wrench}
\end{equation}
is the cable-generated payload wrench.

\subsection{Redundancy analysis}

For a prescribed payload trajectory, the nominal model determines the required
wrench \(\bm{w}\) through \eqref{eq:required_payload_wrench}. The cable forces
\(\bm{f}_i=T_i\bm{s}_i\) must reproduce this wrench through
\eqref{eq:wrench_mapping_matrix}, while satisfying the tension, actuation,
collision-avoidance, cable-interference, and kinematic constraints.

To analyze tension-distribution redundancy, the cable directions are held
fixed and only their magnitudes are treated as unknown. Substituting
\(\bm{f}_i=T_i\bm{s}_i\) into the complete force-and-moment balance
in \eqref{eq:wrench_mapping_matrix} gives
\begin{equation}
\mathbf{A}\bm{T}=\bm{w},
\quad
\mathbf{A}=
\begin{bmatrix}
\bm{s}_1 & \cdots & \bm{s}_n\\
[\mathbf{R}\bm{\rho}_1]_{\times}\bm{s}_1 & \cdots &
[\mathbf{R}\bm{\rho}_n]_{\times}\bm{s}_n
\end{bmatrix},
\label{eq:tension_wrench_mapping}
\end{equation}
where \(\bm{T}=[T_1,\ldots,T_n]^{\mathsf{T}}\). The tension solution
is unique when \(\mathbf{A}\) has full column rank. 
To quantify the proximity
to rank deficiency, we
define
\begin{equation}
\begin{aligned}
k({\mathbf{A}})
&=
\sqrt{
\frac{\lambda_{\min}
\!\left(\mathbf{A}^{\mathsf{T}}\mathbf{A}\right)}
{\lambda_{\max}
\!\left(\mathbf{A}^{\mathsf{T}}\mathbf{A}\right)}}
=\kappa_2^{-1}(\mathbf{A})\in[0,1].
\end{aligned}
\label{eq:tension_allocation_inverse_condition_number}
\end{equation}
The inverse condition number \(k\) provides a normalized measure of the
health of the tension-allocation matrix. A value close to one indicates a
well-conditioned matrix, whereas \(k\) approaches zero as the columns
become linearly dependent. It is exactly zero when
\(\operatorname{rank}(\mathbf{A})<n\), indicating tension redundancy. A small
positive value identifies a near-degenerate configuration in which small wrench
or geometric errors can cause large changes in the allocated tensions. 

A representative degenerate configuration has attachment points distributed
symmetrically about the payload center of mass and cables with the same
inclination. For example, with \(\mathbf{R}=\mathbf{I}_3\), define the \(i\)th
attachment point and cable direction as
\begin{equation}
\begin{aligned}
\bm{\rho}_i&=r
\begin{bmatrix}\cos\theta_i&\sin\theta_i&0\end{bmatrix}^{\mathsf{T}},
\\
\bm{s}_i&=
\begin{bmatrix}
\sin\alpha\cos\theta_i&
\sin\alpha\sin\theta_i&
\cos\alpha
\end{bmatrix}^{\mathsf{T}}.
\end{aligned}
\label{eq:symmetric_cable_configuration}
\end{equation}
Then
\([\bm{\rho}_i]_{\times}\bm{s}_i
=r\cos\alpha[\sin\theta_i,-\cos\theta_i,0]^{\mathsf{T}}\), so every column of
\(\mathbf{A}\) is a linear combination of terms in
\(1\), \(\cos\theta_i\), and \(\sin\theta_i\). Consequently,
\(\operatorname{rank}(\mathbf{A})\leq3\), and any such configuration with
\(n\geq4\) has a nontrivial tension null space and
\(k(\mathbf{A})=0\). Small departures from perfect symmetry
generally produce a small positive inverse condition number, so the force
distribution remains highly sensitive. In contrast, for three UAVs, full
column rank and an inverse condition number bounded away from zero are obtained
generically by choosing non-collinear attachment points, so degeneracy is
straightforward to avoid.

In the degenerate four-or-more-UAV configurations considered here, model
mismatch and low-level tracking errors can shift the realized tensions along a
null-space direction while preserving the payload wrench; in near-degenerate
configurations, the same errors are amplified as the inverse condition number
approaches zero. The problem is therefore to track the commanded payload pose, select and
realize a feasible tension distribution, and preserve the agility of the
existing trajectory-based approach.

\section{Methodology}

\begin{figure*}[t]
    \centering
    \includegraphics[width=0.8\textwidth]{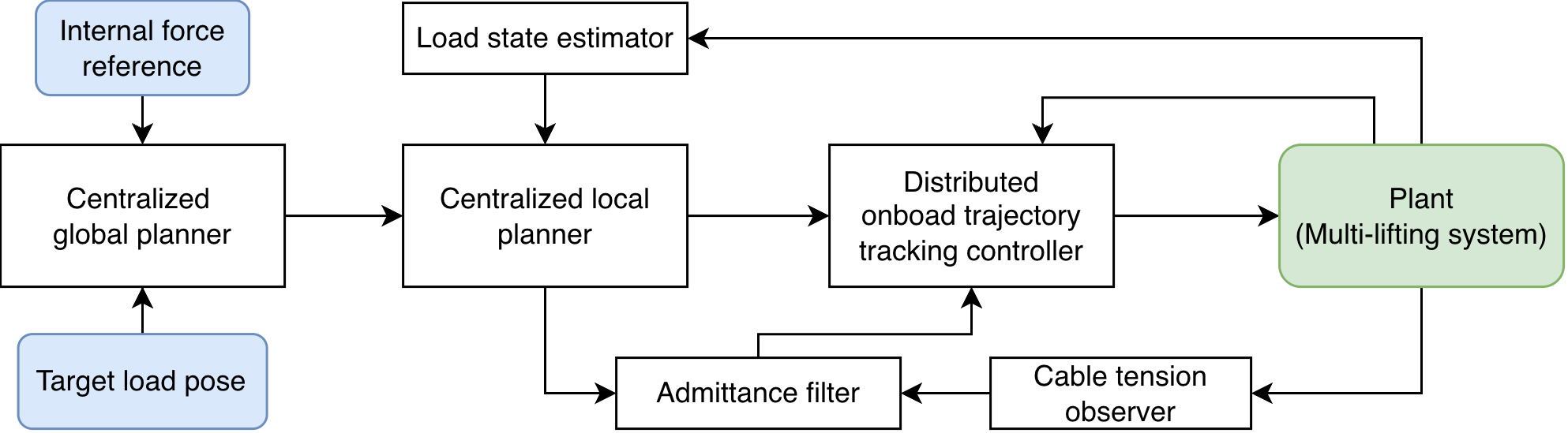}
\caption{Information flow of the proposed framework. The centralized global planner provides payload and cable references to the NMPC. Using the load estimator's feedback the NMPC computes dynamically feasible system trajectories, which are then mapped to individual quadrotor trajectory references. These references are adjusted by the admittance filter using the estimated cable tensions before being tracked by the distributed onboard trajectory controllers.}
    \label{fig:systemOverview}
\end{figure*}

The proposed framework augments the trajectory-based architecture of
\cite{Sun2025AgileCable} with explicit force-distribution planning and local
force feedback. As summarized in Fig.~\ref{fig:systemOverview}, an
internal-force-aware global planner first generates a reference trajectory for
the payload and cables. A centralized NMPC local planner then converts these
references into dynamically feasible, constraint-satisfying trajectories for
the complete multi-lifting system. Relative to \cite{Sun2025AgileCable}, the
system model, constraints, solver, and trajectory decoder are retained, while
the NMPC objective is augmented with the cable-force and cable-direction
references supplied by the global planner. Finally, an admittance filter uses
the discrepancy between the planned and observed cable tensions to modify the
kinematic reference of each quadrotor. This hybrid motion--force structure
preserves the agility of trajectory-based control while regulating the force
distribution that can otherwise become indeterminate for systems with four or
more quadrotors.

\subsection{Internal-force-aware global planner}

Given the current and target payload poses and a user-specified internal-force
level, the global planner constructs the payload and cable reference

\begin{equation}
\mathcal{T}^{g}
=
\left\{
\bm{p}^{g},\mathbf{R}^{g},\bm{v}^{g},\bm{\omega}^{g},
\left\{\bm{s}_{i}^{g},T_i^{g}\right\}_{i=1}^{n}
\right\}.
\label{eq:global_reference}
\end{equation}

The payload position is generated using a minimum-snap polynomial
\cite{Mellinger2011MinimumSnap}, while its orientation is interpolated using
SLERP \cite{Shoemake1985QuaternionCurves}. Their derivatives provide
\(\bm{v}^{g}\) and \(\bm{\omega}^{g}\), and the nominal payload model gives the
required wrench \(\bm{w}^{g}\). Following
\cite{Lee2018Geometric,Geng2022LoadDistribution}, let \(\mathbf{Z}^{g}\) be an
orthonormal null-space basis of \(\mathbf{P}(\mathbf{R}^{g})\). Wrench-preserving
force allocations are parameterized by

\begin{equation}
\begin{aligned}
\bm{f}_{\min}^{g}
&=\left[\mathbf{P}(\mathbf{R}^{g})\right]^{\dagger}\bm{w}^{g},\\
\bm{f}^{g}(\bm{\lambda})
&=\bm{f}_{\min}^{g}+\mathbf{Z}^{g}\bm{\lambda},
\qquad
\mathbf{P}(\mathbf{R}^{g})\mathbf{Z}^{g}=\mathbf{0}.
\end{aligned}
\label{eq:nullspace_force}
\end{equation}

The coefficient \(\bm{\lambda}\) is chosen to maintain cable tautness and impose
a user-specified outward preload. For
\(\bm{\rho}_i=[x_i,y_i,z_i]^{\mathsf{T}}\), define the outward direction
\(\bm{r}_i^{g}=\mathbf{R}^{g}[x_i,y_i,0]^{\mathsf{T}}/
\sqrt{x_i^2+y_i^2}\). The stacked preload target is
\(\bm{f}_{\mathrm{rad}}^{g}=F_{\mathrm{int}}
[(\bm{r}_1^{g})^{\mathsf{T}},\ldots,
(\bm{r}_n^{g})^{\mathsf{T}}]^{\mathsf{T}}\), with null-space projection
\(\bm{\lambda}_{\mathrm{rad}}^{g}=(\mathbf{Z}^{g})^{\mathsf{T}}
\bm{f}_{\mathrm{rad}}^{g}\). Let \(\bm{f}_i^{g}(\bm{\lambda})\) denote the
\(i\)th three-dimensional block of \eqref{eq:nullspace_force}. The allocation is
obtained from

\begin{equation}
\begin{aligned}
\min_{\bm{\lambda}}\quad
J_g(\bm{\lambda})
={}&
\frac{k_r}{2}
\left\|\bm{\lambda}-\bm{\lambda}_{\mathrm{rad}}^{g}\right\|^2
\\
&+\frac{k_f}{2}
\sum_{i=1}^{n}
\left[
\max\!\left(0,T_{\min}-\left\|\bm{f}_i^{g}(\bm{\lambda})\right\|\right)
\right]^2
\\
&+\frac{k_n}{2}
\sum_{i=1}^{n}
\left[
\max\!\left(0,F_{\mathrm{int,min}}
-F_{r,i}^{g}(\bm{\lambda})\right)
\right]^2 ,
\end{aligned}
\label{eq:null_space_optimization}
\end{equation}
where \(F_{r,i}^{g}=(\bm{r}_i^{g})^{\mathsf{T}}
\bm{f}_i^{g}(\bm{\lambda})\), \(k_r,k_f,k_n>0\), and
\(F_{\mathrm{int,min}}\) is the minimum outward force component. The optimized
forces are converted into \(T_i^{g}=\|\bm{f}_i^{g}\|\) and
\(\bm{s}_i^{g}=\bm{f}_i^{g}/T_i^{g}\), completing
\eqref{eq:global_reference} while preserving the required payload wrench.

\subsection{Force-reference-aware NMPC local planner}

The local planner follows the receding-horizon NMPC formulation introduced in
our previous work \cite{Sun2025AgileCable}. We briefly state the formulation to
clarify the modification made in this work. At each planning update, the
estimated load--cable state initializes a finite-horizon optimal control
problem. Defining the decision matrices
\(\mathbf{X}=[\bm{x}_0,\ldots,\bm{x}_N]\) and
\(\mathbf{U}=[\bm{u}_0,\ldots,\bm{u}_{N-1}]\), the problem is

\begin{equation}
\begin{aligned}
\min_{\mathbf{X},\mathbf{U}}\quad
J_{\mathrm{N}}
={}&
\sum_{k=0}^{N-1}\ell_k(\bm{x}_k,\bm{u}_k)
+\ell_N(\bm{x}_N),
\\
\mathrm{s.t.}\quad
&\bm{x}_0=\hat{\bm{x}},
\qquad
\bm{x}_{k+1}=\mathcal{F}(\bm{x}_k,\bm{u}_k),
\\
&\bm{h}(\bm{x}_k,\bm{u}_k)\leq\bm{0},
\qquad k=0,\ldots,N-1,
\end{aligned}
\label{eq:nmpc_ocp}
\end{equation}
where \(\mathcal{F}\) is the discrete-time state-transition model obtained from
the coupled cable--load model: the cable kinematics in
\eqref{eq:cable_kinematics} and the payload dynamics in
\eqref{eq:payload_dynamics}. Thus, the NMPC prediction uses the same physical
model introduced in the preliminaries. The vector \(\bm{h}\) collects the path
constraints, including cable tautness, actuation limits, and
collision-avoidance constraints. The state contains the payload pose and twist
together with the cable directions, tensions, and their required derivatives.
As in \cite{Sun2025AgileCable}, a decoder maps the optimized load--cable
trajectory to position, velocity, acceleration, attitude, and feedforward
references for every quadrotor.

Here, \(J_{\mathrm{N}}\) is the finite-horizon objective,
\(\ell_k\) and \(\ell_N\) are the stage and terminal costs, and \(N\) is the
prediction-horizon length. The base planner promotes payload tracking and
smooth control but does not
explicitly select the redundant force distribution generated by the global
planner. In this work, its stage cost is augmented as

\begin{equation}
\begin{aligned}
\ell_k
={}\ell_k^{\mathrm{base}}
+\sum_{i=1}^{n}
\left\|\bm{s}_{i,k}-\bm{s}_{i,k}^{g}\right\|_{\mathbf{Q}_s}^{2} +\sum_{i=1}^{n}
\left\|
T_{i,k}\bm{s}_{i,k}-\bm{f}_{i,k}^{g}
\right\|_{\mathbf{Q}_f}^{2}
\end{aligned}
\label{eq:force_aware_nmpc_cost}
\end{equation}
with analogous terminal penalties. Here,
\(\ell_k^{\mathrm{base}}\) denotes the payload-state and control-effort
costs, that minimizes the pose and twist tracking errors, while \(\mathbf{Q}_s\) and \(\mathbf{Q}_f\) weight cable-direction and
cable-force tracking, respectively. Thus, the global planner chooses a desired
force allocation, whereas the NMPC reconciles that allocation with the coupled
dynamics and constraints over the prediction horizon. 

Penalizing force error in the NMPC improves the planned distribution, but the
optimized force is still realized through kinematic quadrotor trajectories and
is not a directly actuated output. The NMPC therefore cannot by itself reject
the difference between the nominal force allocation and the tension distribution
realized by the physical system. This motivates the force-feedback layer below.

\subsection{Admittance filter for hybrid cable-tension and trajectory tracking}

The admittance filter specifically addresses degradation of the tension
allocation characterized by \eqref{eq:tension_allocation_inverse_condition_number}.
When \(k({\mathbf{A}})\) approaches zero, small attachment-geometry,
cable-length, or trajectory-tracking errors can produce large tension
redistributions while causing little change in the resultant payload wrench.
The filter closes this otherwise unregulated internal-force channel by using
the observed tension error to correct each quadrotor's kinematic reference.

For each quadrotor, the local planner provides a commanded tension
\(T_{i,\mathrm{cmd}}\), a cable direction \(\bm{s}_{i,\mathrm{cmd}}\), and a
kinematic trajectory
\((\bm{p}_{i,\mathrm{cmd}},\bm{v}_{i,\mathrm{cmd}},
\bm{a}_{i,\mathrm{cmd}})\). The onboard wrench observer provides an external
force estimate \(\bm{f}_{i,\mathrm{ext}}\). 
The force estimate is obtained as such:
\begin{equation}
\mathbf{f}_{i,ext} = m_d\,LPF(\mathbf{a}^B_{IMU} - \mathbf{b}_a) - \mathbf{f}^B_{drag} -
\begin{bmatrix}
0 \\
0 \\
T
\end{bmatrix},
\label{eq:force_observer}
\end{equation}
where \(m_d\) is the quadrotor mass, \(LPF(\cdot)\) denotes a low-pass filter, \(\mathbf{a}^B_{IMU}\) is the acceleration measured by the IMU expressed in the body frame \(B\), \(\mathbf{b}_a\) is the accelerometer bias, \(\mathbf{f}^B_{drag}\) is the modeled aerodynamic drag force expressed in the body frame, and \(T\) is the total thrust generated by the quadrotor.

Using the implemented sign
convention, the observed tension is obtained by projection onto the commanded
cable direction,

\begin{equation}
T_{i,\mathrm{obs}}
=
-\bm{f}_{i,\mathrm{ext}}^{\mathsf{T}}\bm{s}_{i,\mathrm{cmd}}.
\label{eq:tension_observer}
\end{equation}

The tension error is converted into an equivalent correction force aligned
with the cable,

\begin{equation}
e_{T,i}
=
T_{i,\mathrm{cmd}}-T_{i,\mathrm{obs}},
\qquad
\bm{f}_{a,i}
=
e_{T,i}\bm{s}_{i,\mathrm{cmd}}.
\label{eq:tension_error}
\end{equation}

This force drives a virtual mass--spring--damper system,

\begin{equation}
\mathbf{M}_{a,i}\dot{\bm{v}}_{a,i}
+\mathbf{D}_{a,i}\bm{v}_{a,i}
+\mathbf{K}_{a,i}\bm{p}_{a,i}
=
\bm{f}_{a,i},
\qquad
\dot{\bm{p}}_{a,i}=\bm{v}_{a,i},
\label{eq:admittance_dynamics}
\end{equation}
where \(\mathbf{M}_{a,i}\), \(\mathbf{D}_{a,i}\), and
\(\mathbf{K}_{a,i}\) are the virtual inertia, damping, and stiffness. The
resulting corrections modify the NMPC-generated kinematic reference,

\begin{equation}
\begin{aligned}
\bm{p}_{i,\mathrm{ref}} &= \bm{p}_{i,\mathrm{cmd}}+\bm{p}_{a,i},
&\quad
\bm{v}_{i,\mathrm{ref}} &= \bm{v}_{i,\mathrm{cmd}}+\bm{v}_{a,i},\\
\bm{a}_{i,\mathrm{ref}} &= \bm{a}_{i,\mathrm{cmd}}+\dot{\bm{v}}_{a,i}.
\end{aligned}
\label{eq:admittance_reference}
\end{equation}

Consequently, the NMPC remains responsible for coordinated whole-body motion
and constraint satisfaction, while the admittance filter adds a local compliant
motion that reduces the discrepancy between predicted and realized cable
tension. The filter does not alter the global payload command; it only adjusts
the individual quadrotor references before they are tracked by the onboard
controller. Section~\ref{sec:cable_tension_redundancy} demonstrates the resulting improvement
from the admittance filter when the tension-allocation matrix is degenerate or
nearly degenerate, corresponding to
\(k({\mathbf{A}})=0\) or \(k({\mathbf{A}})\approx0\),
respectively.

\subsection{Onboard trajectory tracking}

The corrected references are tracked by the same flatness-based PD and INDI
controller used in \cite{Sun2025AgileCable}. The flatness mapping generates the
required attitude and feedforward quantities, while INDI rejects unmodeled force
and torque disturbances using onboard inertial measurements
\cite{tal_accurate_2021,smeur_cascaded_2018}. This component is unchanged in the
present work; its role is to execute the admittance-corrected local trajectory
and generate the rotor commands.
\section{Results}
We evaluated the proposed approach in simulation and real-world experiments.
The evaluation addresses the following questions:
\begin{itemize}
    \item Can the proposed null-space optimization generate cable-tension references that satisfy the required tension constraints?
    
    \item Can the admittance filter improve cable-tension tracking and produce a more uniform load distribution among the quadrotors?
    
    \item How do the proposed modifications affect payload trajectory tracking and robustness to model mismatch?
\end{itemize}
\subsection{Experimental setup}
Simulations were conducted in RotorS \cite{Furrer2016}, and real-world
experiments were performed in a Vicon-equipped indoor arena. The centralized
planner ran at 10~Hz on an external computer. Each quadrotor ran the Agilicious
flight stack \cite{foehn2022agilicious} with onboard control at 300~Hz. Vicon
measurements were provided at 100~Hz and fused with onboard sensors using an
extended Kalman filter. Unless stated otherwise, the experiments used four
quadrotors and \SI{1}{\meter} cables; larger teams were used to assess
scalability. The parameters in Table~\ref{tab:system_parameters} were used
throughout. Hovering experiments were compared with the geometric controller in
\cite{Lee2018Geometric} and the trajectory-based framework in
\cite{Sun2025AgileCable}; trajectory-tracking experiments used the latter as the
baseline.
\begin{table}[h]
    \caption{System Parameters}
    \label{tab:system_parameters}
    \centering
    \footnotesize
    \setlength{\tabcolsep}{4pt}
    \begin{tabular}{lclc}
        \hline
        \textbf{Parameter} & \textbf{Value} &
        \textbf{Parameter} & \textbf{Value} \\
        \hline
        $m_L$ (4 UAVs)  & 1.4 kg  & $m_L$ (10 UAVs) & 4 kg \\
        $m_i$ (4 UAVs)  & 0.65 kg & $m_i$ (10 UAVs) & 0.6 kg \\
        $l$              & 1 m     & $K_a$            & 60 N/m \\
        $D_a$            & 50 Ns/m & $M_a$            & 10 kg \\
        $k_r$            & 1       & $k_f$             & 2.5 \\
        $k_n$            & 7       & $Q_t$             & 1.0 \\
        $Q_s$            & $[100,\,100,\,100]$ & & \\
        \hline
        $|\bm{\rho}_i|$, 4 UAVs
        & $\begin{bmatrix}0.24&0.22&0.16\end{bmatrix}^{T}$ m
        & \multicolumn{2}{l}{$i=1,\ldots,4$} \\[1mm]

        $|\bm{\rho}_i|$, 10 UAVs
        & $\begin{bmatrix}2&0.5&0.1\end{bmatrix}^{T}$ m
        & \multicolumn{2}{l}{$i=1,5,6,10$} \\

        & $\begin{bmatrix}1&0.5&0.1\end{bmatrix}^{T}$ m
        & \multicolumn{2}{l}{$i=2,4,7,9$} \\

        & $\begin{bmatrix}0&0.5&0.1\end{bmatrix}^{T}$ m
        & \multicolumn{2}{l}{$i=3,8$} \\
        \hline
    \end{tabular}
\end{table}

\subsection{Simulation results}
\subsubsection{Null-space optimization validation}
\begin{figure}[t]
    \centering
    \includegraphics[width=0.9\linewidth]{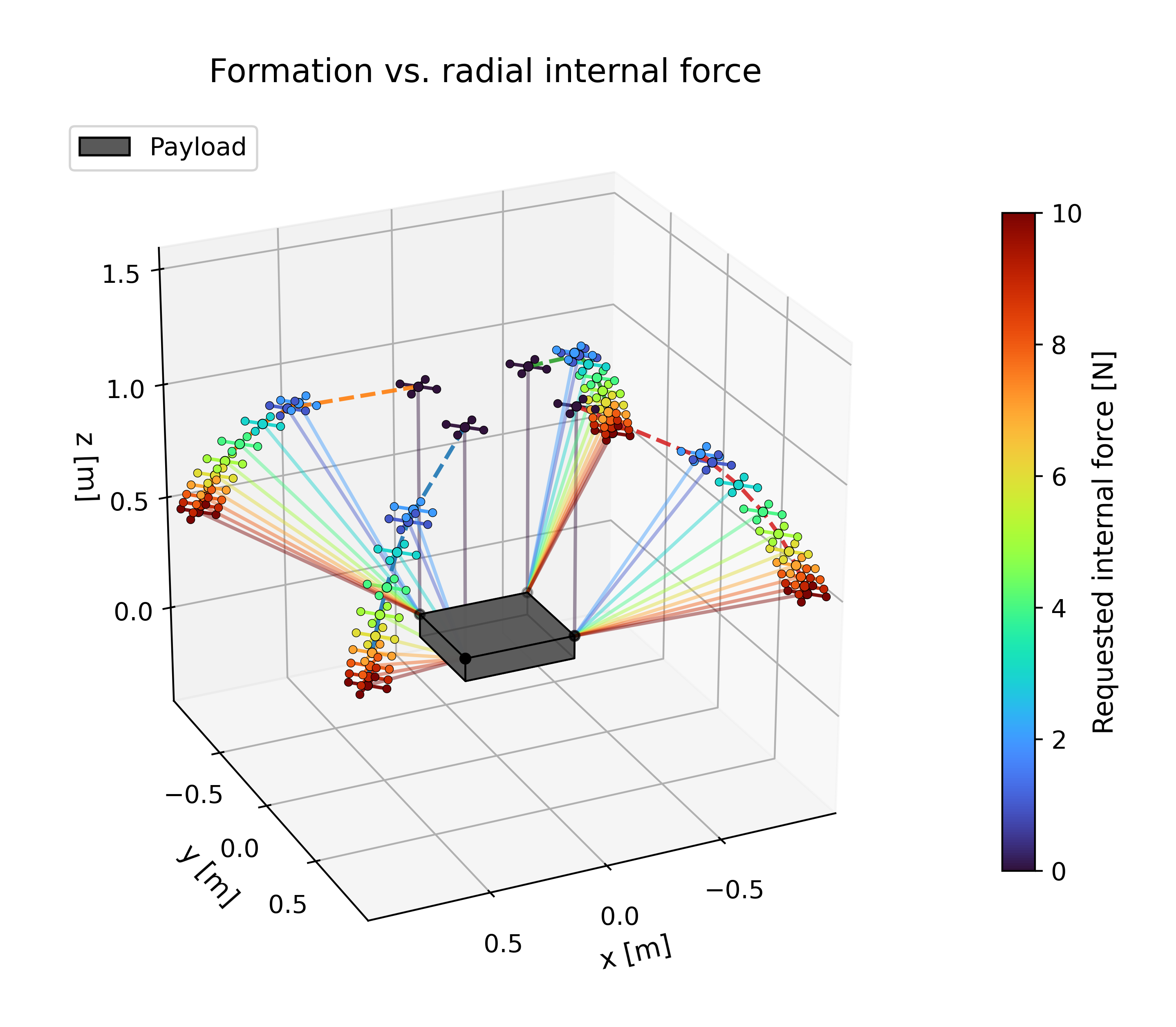}
    \caption{Effect of the radial preload magnitude \(F_{\mathrm{int,min}}\) on the
    quadrotor formation.}
    \label{fig:flatForm}
\end{figure}

Figure~\ref{fig:flatForm} illustrates how the radial preload changes the
resulting formation.

\begin{figure}[h]
    \centering
    \includegraphics[width=1\linewidth]{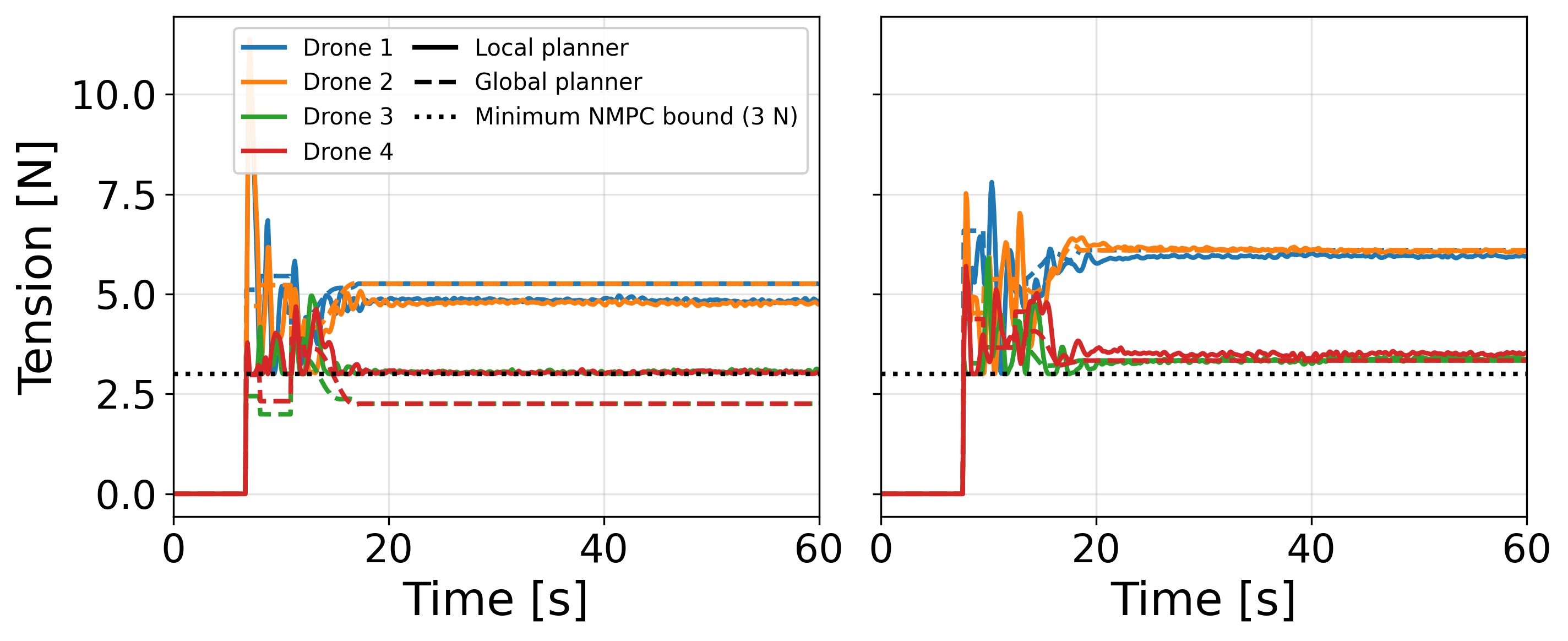}
    \caption{Tension tracking without and with null-space optimization.}
    \label{fig:globalVsLocal}
\end{figure}

A four-quadrotor hover with a payload pitch of \(30^\circ\) was then used to
evaluate the force-allocation strategy. Without null-space optimization, the
global planner generated references below the local planner's minimum tension
of \(3\,\mathrm{N}\). As shown in Fig.~\ref{fig:globalVsLocal}, enforcing the
tension constraints through null-space optimization substantially reduces the
tracking error by providing the local planner with feasible references.




\subsubsection{Cable-tension redundancy}
\label{sec:cable_tension_redundancy}
We investigated the force-allocation characteristics of four- and ten-UAV
configurations. The UAVs were distributed symmetrically around the payload, so
particular payload orientations produced nearly degenerate allocation matrices.
The payload hovered in level and inclined orientations, with and without the
admittance filter.

For each case, \(\mathbf{A}\) was computed from the payload orientation
\(\mathbf{R}\) and cable directions \(\bm{s}_i\). Changing the payload
inclination therefore changed the conditioning of \(\mathbf{A}\). For four
UAVs, the inverse condition number \(k(\mathbf{A})\) was computed using
\eqref{eq:tension_allocation_inverse_condition_number}. For ten UAVs,
\(\mathbf{A}\) had more columns than rows and therefore admitted infinitely
many tension solutions.

\begin{figure}
    \centering
    \includegraphics[width=1.0\linewidth]{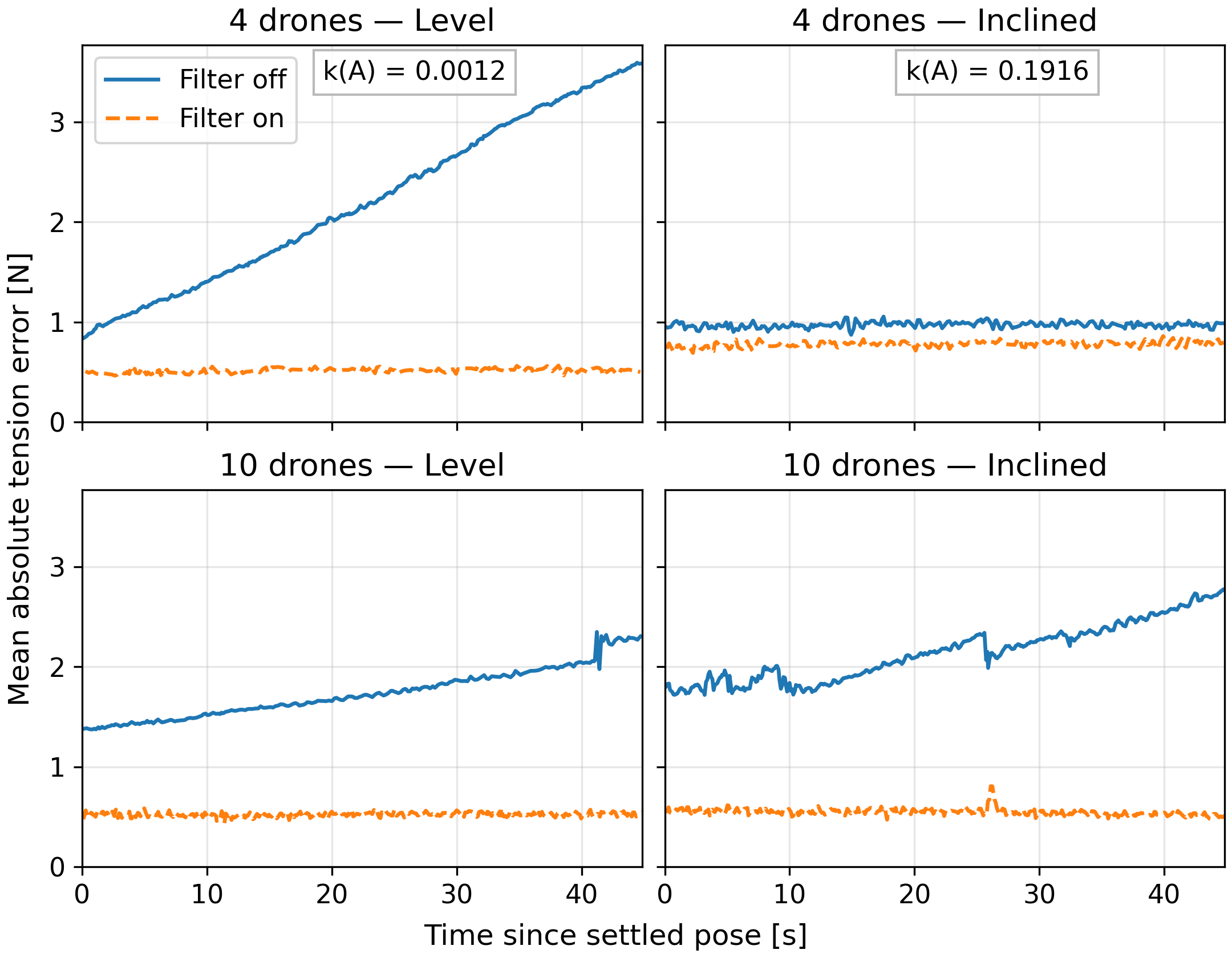}
    \caption{Effect of the allocation-matrix conditioning on cable-tension tracking for the four- and ten-drone configurations, with and without the admittance filter.}
    \label{fig:redundancy}
\end{figure}

Figure~\ref{fig:redundancy} presents the mean absolute tension-tracking error.
For the level four-UAV configuration, \(k(\mathbf{A})=0.0012\), indicating a
poorly conditioned allocation. The resulting redundancy increased the tension
error when the filter was disabled, whereas activating the filter substantially
reduced it. In the inclined configuration, \(k(\mathbf{A})=0.1916\), and the
tension error remained low even without the filter.

For ten UAVs, \(\mathbf{A}\) lacked full column rank for every cable geometry.
Consequently, the mean tension error grew in both level and inclined
configurations when the admittance filter was disabled.
\subsubsection{Effect of the admittance filter}





\begin{figure}[t]
    \centering
    \begin{subfigure}[b]{1\linewidth}
        \centering
        \includegraphics[width=\linewidth]{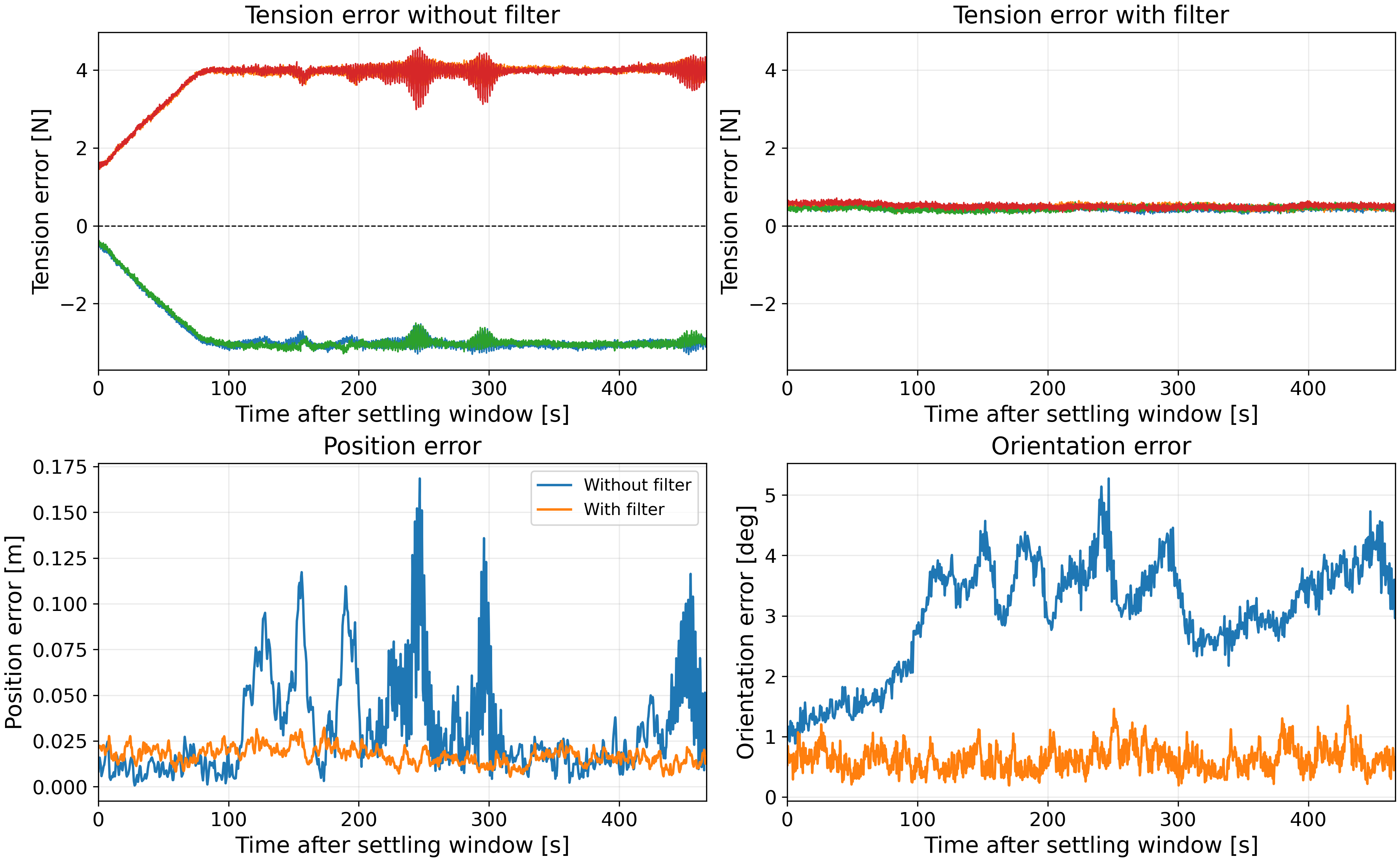}
        \caption{Four-UAV configuration.}
        \label{fig:4dronessim}
    \end{subfigure}

    \par\vspace{0.3cm}

    \begin{subfigure}[b]{1\linewidth}
        \centering
        \includegraphics[width=\linewidth]{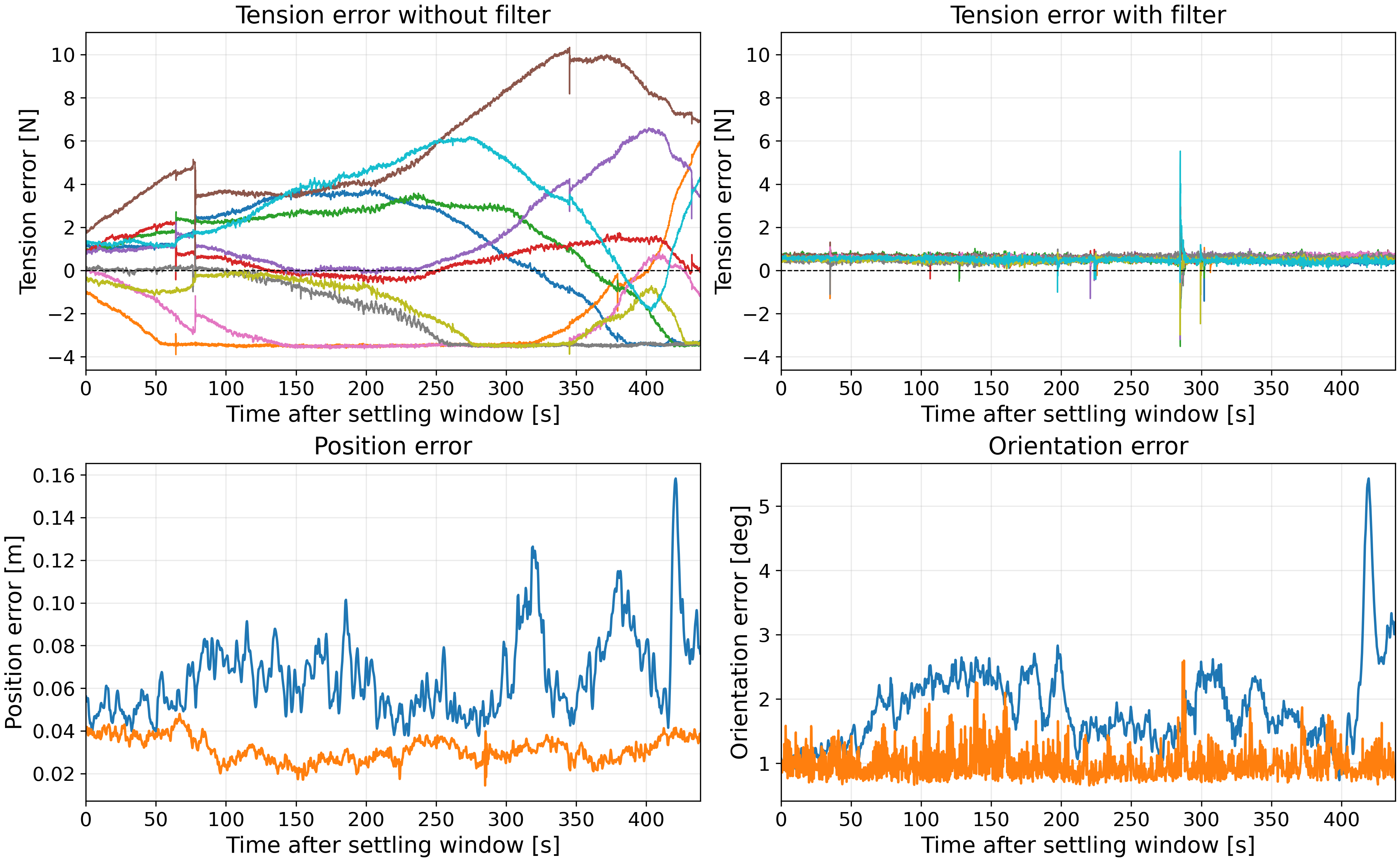}
        \caption{Ten-UAV configuration.}
        \label{fig:10dronesim}
    \end{subfigure}

    \caption{Tension- and payload-tracking performance with and without the
    admittance filter.}
    \label{fig:comparison10}
\end{figure}

\begin{figure}[h]
    \centering
    \includegraphics[width=1.0\linewidth]{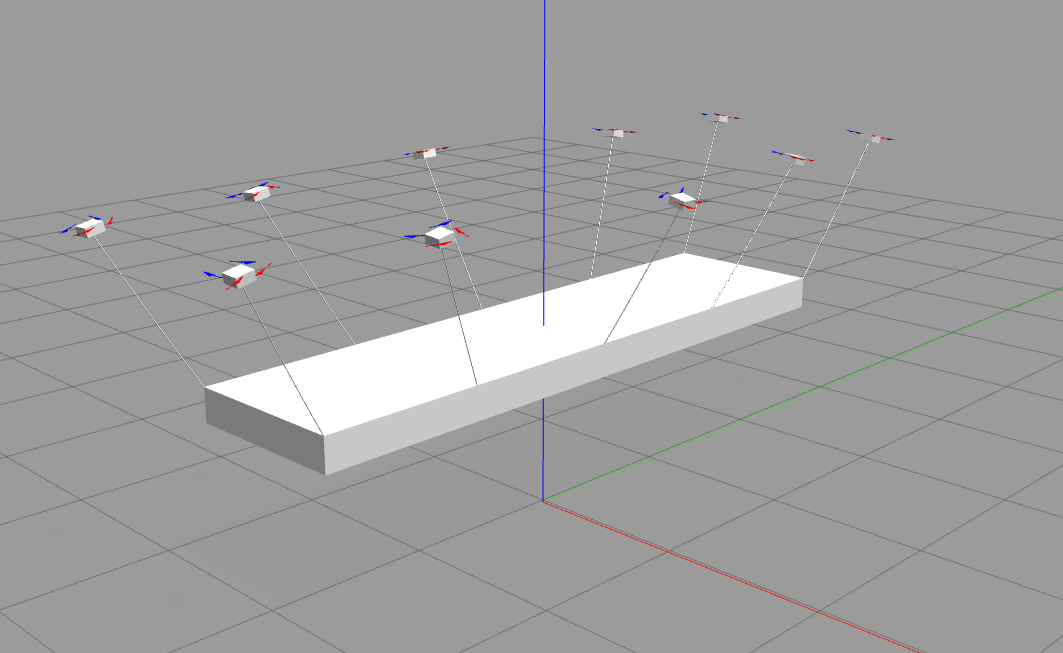}
    \caption{Ten UAVs lifting a payload in the simulated level configuration.}
    \label{fig:10dronesliftsim}
\end{figure}

Stationary hover was first used to evaluate tension regulation under nominal
conditions because the symmetric geometry yielded small \(k(\mathbf{A})\).
Figure~\ref{fig:comparison10} compares performance with and without the
admittance filter for four- and ten-quadrotor configurations.
Without the filter, noticeable tension imbalance remains despite the absence of
model mismatch. With the filter enabled, the measured tensions converge closer
to their references and the spread among cables is reduced. The same
decentralized correction law remained effective for ten quadrotors,
demonstrating scalability to larger teams. The filter also reduced the payload
position and orientation errors. Figure~\ref{fig:10dronesliftsim} shows the
ten-UAV simulation.

\subsubsection{Sensitivity to inertial model uncertainties}
Because the force references depend on the payload model, we varied the payload
mass and inertia assumed by the planner while holding the simulated system
fixed. The proposed framework, with and without the admittance filter, was
compared with the geometric controller in \cite{Lee2018Geometric}.

\begin{figure}[t]
    \centering
    \includegraphics[width=\linewidth]{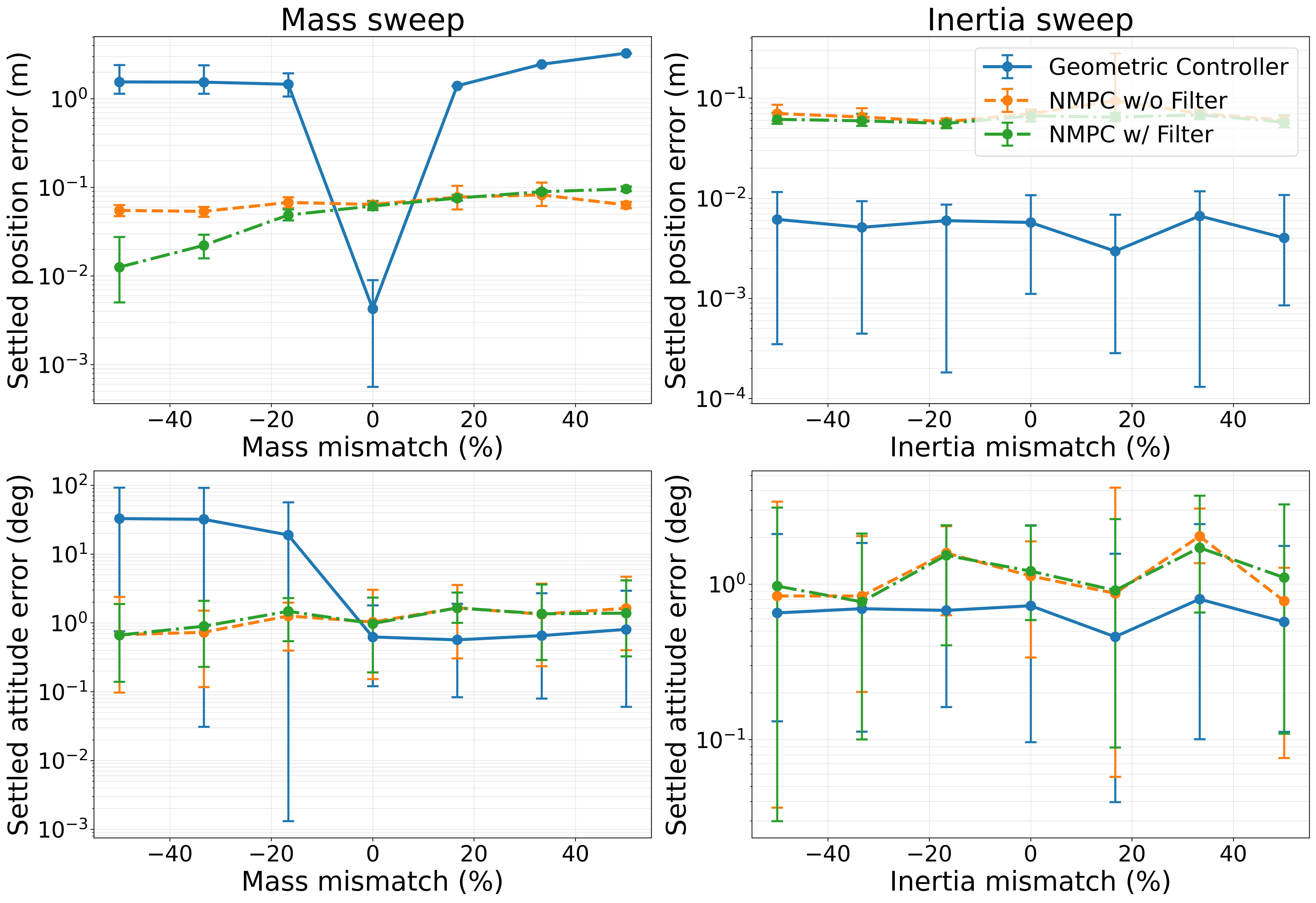}
    \caption{Sensitivity of the geometric controller and the NMPC-based
    framework to payload mass and inertia mismatch during stationary hovering.}
    \label{fig:nmpcvslee}
\end{figure}

Figure~\ref{fig:nmpcvslee} shows that payload-mass mismatch affects the
geometric controller more strongly than inertia mismatch because the controller
uses the payload model to compute the commanded wrench and directly issues force
commands to the UAVs. Both NMPC configurations varied less over the tested range
because the UAVs primarily tracked kinematic references from the local planner.
Notably, adding the admittance filter did not noticeably reduce robustness to inertial-model
uncertainty.

\subsubsection{Agile trajectory tracking}
We next evaluated dynamic performance using a figure-eight trajectory with a
maximum velocity of \SI{5}{\meter\per\second}, acceleration of
\SI{8}{\meter\per\second\squared}, and jerk of
\SI{16}{\meter\per\second\cubed}. The trajectory spanned \SI{5}{\meter} along
the \(x\)-axis and \SI{4}{\meter} along the \(y\)-axis at a constant altitude
of \SI{1}{\meter}.

We also evaluated robustness to cable-length mismatch by introducing a
\SI{-1}{\centi\meter} error in the modeled length of cable~1. As shown in
Table~\ref{tab:sim_trajectory_tracking}, the two configurations achieved
comparable tracking without mismatch because \(k(\mathbf{A})\) was generally
larger during agile motion than during hovering. With the mismatch, the
admittance filter reduced the three-dimensional mean absolute error from
\SI{0.221}{\meter} to \SI{0.199}{\meter}.
\begin{table}[h]
    \centering
    \caption{Payload Trajectory Tracking With and Without Cable-Length Mismatch}
    \label{tab:sim_trajectory_tracking}
    \footnotesize
    \setlength{\tabcolsep}{3pt}
    \begin{tabularx}{\columnwidth}{@{}Xcccc@{}}
        \hline
        \textbf{Mean Absolute Error (m)}
        & \textbf{X} & \textbf{Y} & \textbf{Z} & \textbf{3D} \\
        \hline
        No mismatch, without admittance filter
        & 0.0864 & \textbf{0.1099} & \textbf{0.0439} & \textbf{0.1608} \\

        No mismatch, with admittance filter
        & \textbf{0.0827} & 0.1207 & 0.0455 & 0.1674 \\
        \hline
        1 cm mismatch, without admittance filter
        & 0.092 & 0.090 & 0.132 & 0.221 \\

        1 cm mismatch, with admittance filter
        & \textbf{0.086} & \textbf{0.069} & \textbf{0.129} & \textbf{0.199} \\
        \hline
    \end{tabularx}
\end{table}

\begin{figure}[t!]
    \centering
    \includegraphics[width=1.0\linewidth]{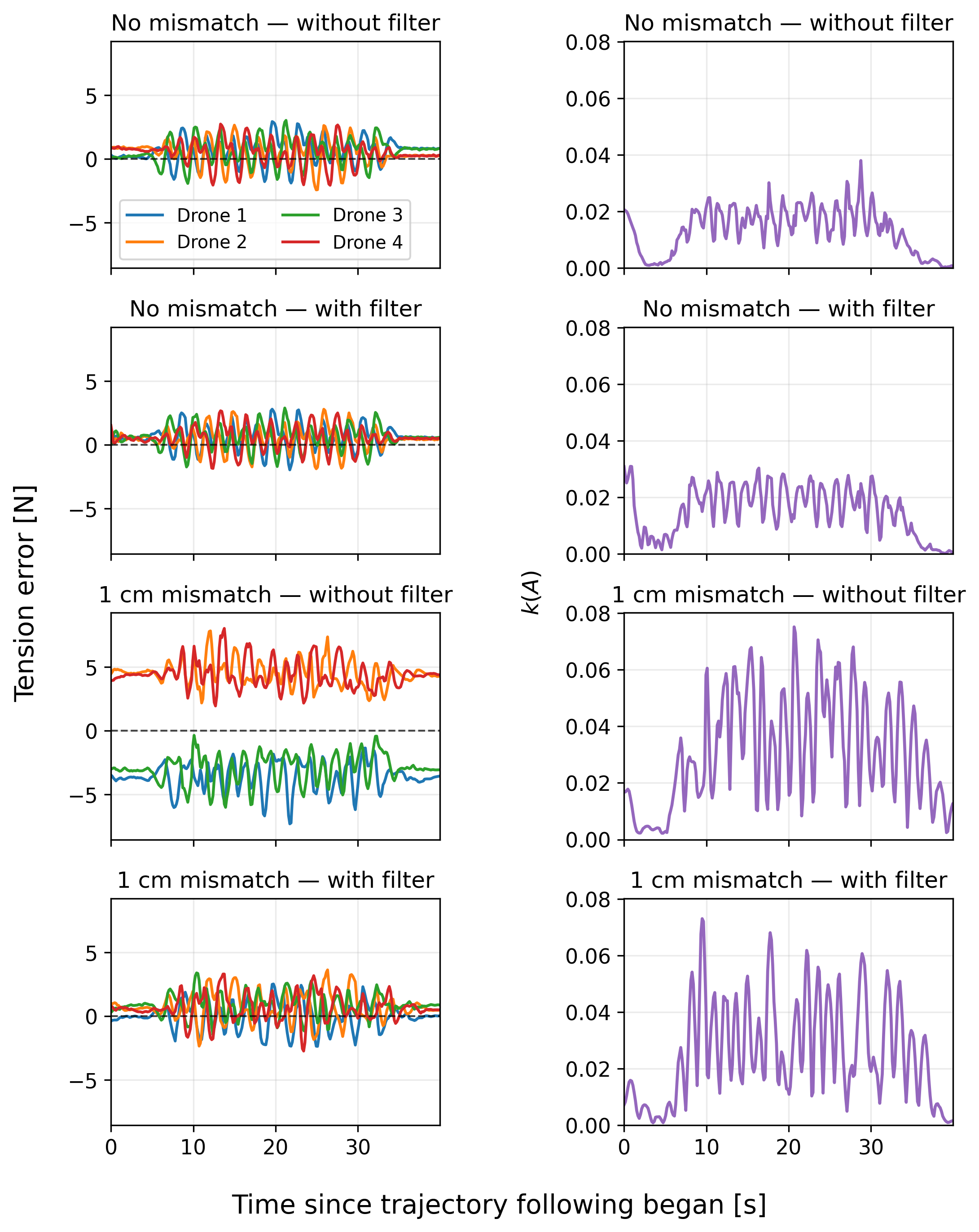}
    \caption{Cable-tension errors and allocation conditioning during agile
    trajectory tracking, with and without cable-length mismatch and admittance
    filtering.}
    \label{fig:comparisonFigure8Mismatch}
\end{figure}

The corresponding cable-tension errors and allocation conditioning are shown in
Fig.~\ref{fig:comparisonFigure8Mismatch}.

The cable-length mismatch produces a more pronounced discrepancy
between the planned and observed cable tensions in the baseline
configuration. With the admittance filter enabled, the individual
quadrotor trajectories are adapted in response to these tension errors,
reducing the discrepancy between the reference and observed forces.
The corresponding reduction in payload tracking error indicates that,
for this case, improving the force distribution also improves the
overall trajectory-tracking performance.

\subsection{Real-world experiments}


\subsubsection{Hovering}
\begin{figure}[t!]
    \centering
    \includegraphics[width=1.0\linewidth]{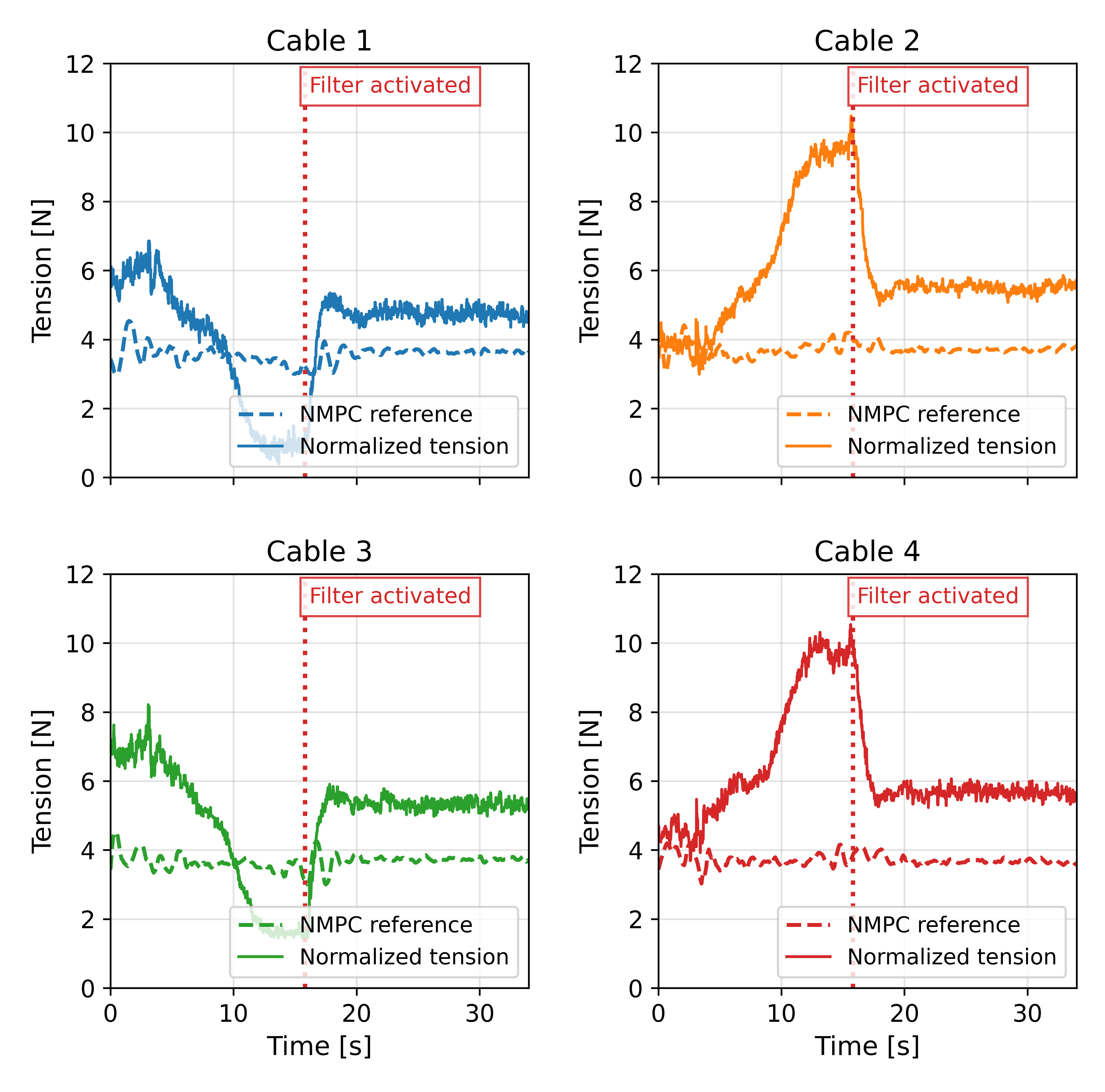}
    \caption{Tension tracking with and without the admittance filter during
    hovering.}
    \label{fig:experimentsHover}
\end{figure}

Geometric mismatch made the force imbalance more pronounced in the real-world
experiments. Figure~\ref{fig:experimentsHover} shows the cable tensions during
hovering. The system initially operated without the admittance filter, which was
activated after approximately \SI{15}{\second}.

Before activation, UAVs~2 and~4, located on diagonally opposite sides of the
payload, carried a larger share of its weight, while the remaining cable
tensions approached zero. After activation, the measured tensions converged
toward their references, producing a more balanced distribution.

A residual steady-state offset of approximately \SI{2}{\newton} remained after
activation, compared with approximately \SI{0.5}{\newton} in simulation
(Fig.~\ref{fig:4dronessim}). This offset may result from unmodeled aerodynamic
effects, particularly rotor downwash on the payload, which are not captured by
the observer in \eqref{eq:force_observer}. Direct tension measurements could
help distinguish observer bias from aerodynamic loading.

\begin{table}[h]
    \centering
    \caption{Payload Trajectory Tracking Performance During Agile Flight}
    \label{tab:experiment_trajectory_tracking}
    \footnotesize
    \setlength{\tabcolsep}{2pt}
    \renewcommand{\arraystretch}{1.0}

    \begin{tabular}{@{}>{\raggedright\arraybackslash}p{0.48\columnwidth}cccc@{}}
        \hline
        \textbf{Mean Absolute Error (m)} 
        & \textbf{X} 
        & \textbf{Y} 
        & \textbf{Z} 
        & \textbf{3D} \\
        \hline

        Baseline \cite{Sun2025AgileCable}
        & 0.0713 & 0.1339 & 0.0506 & 0.1787 \\

        Ours
        & \textbf{0.0711} & \textbf{0.1106} & \textbf{0.0429} & \textbf{0.1525} \\

        Ours (with unknown load)
        & 0.0860 & 0.1339 & 0.0628 & 0.1974 \\

        \hline
    \end{tabular}
\end{table}
\subsubsection{Tracking agile trajectories}

We further perform experiments tracking agile trajectories with the proposed framework, in comparison with the original trajectory-based approach proposed in \cite{Sun2025AgileCable}.
These experiments used the same trajectory as the simulations. As shown in
Table~\ref{tab:experiment_trajectory_tracking}, the proposed framework achieved
accurate tracking during agile motion. We also introduced an unmodeled sloshing
load by placing a basketball (0.6kg) inside the payload, thereby applying additional
force and torque disturbances (Fig.~\ref{fig:eye-catcher}).

\begin{figure}[t!]
    \centering
    \includegraphics[width=1.0\linewidth]{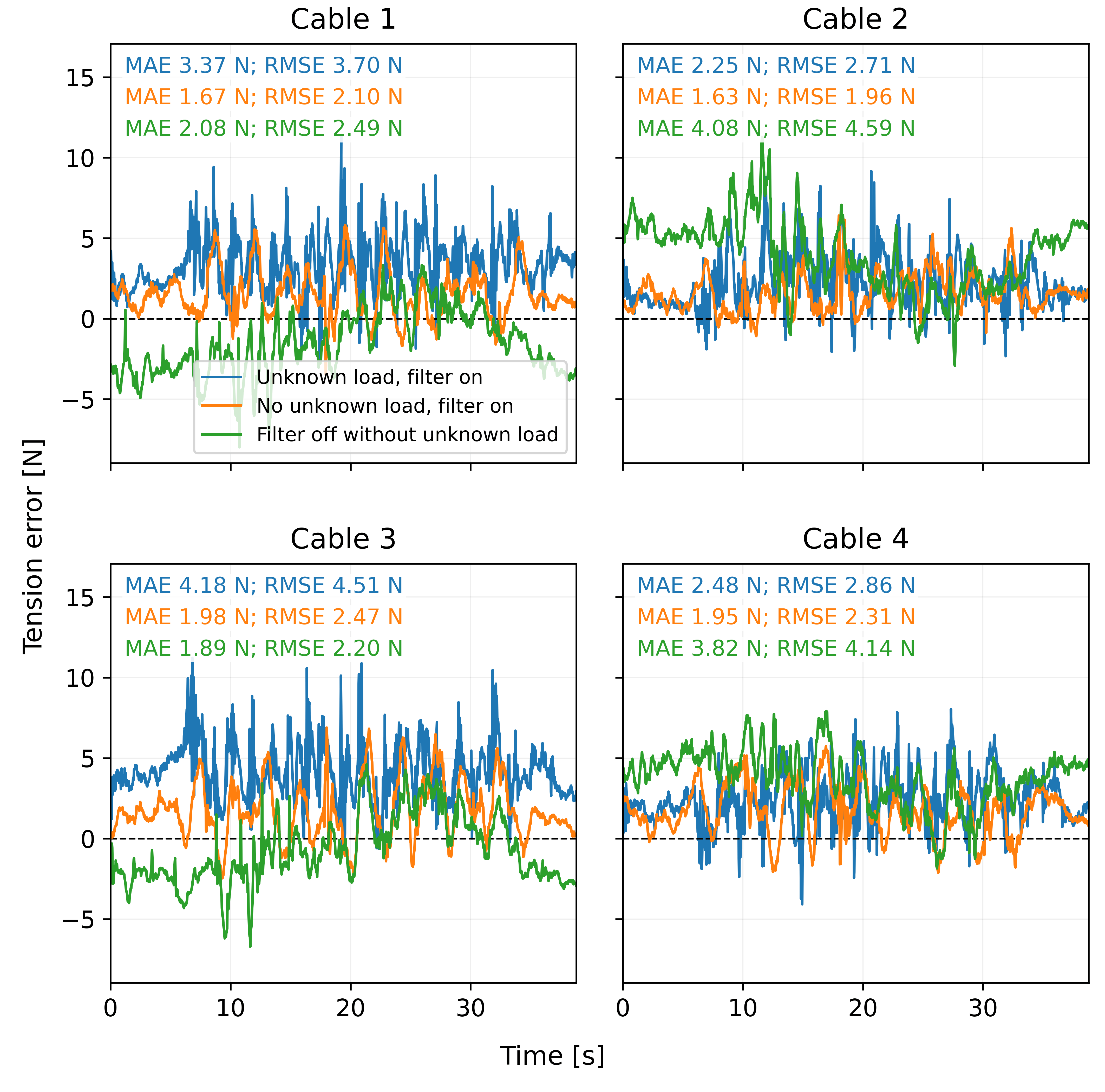}
    \caption{Cable-tension errors during agile maneuvering.}
    \label{fig:cable_tension_tracking}
\end{figure}

Figure~\ref{fig:cable_tension_tracking} compares the tension-tracking errors of
the three configurations. The proposed framework without the unknown load had
the lowest mean absolute error for all four cables. With the unknown load, it
still outperformed the baseline on cables~1, 3, and~4, although the error
increased on cable~2. Across all four cables, enabling the admittance filter
reduced the aggregate tension-tracking RMSE from \SI{3.51}{\newton} to
\SI{2.09}{\newton}.

\section{Conclusions}
This paper presented a hybrid planning-and-control framework for improving
cable-force distribution in trajectory-based aerial multi-lifting systems with
four or more UAVs. An internal-force-aware global planner exploits the null
space of the payload wrench-allocation matrix to generate feasible cable-force
and direction references, which are incorporated into the cost of a
centralized NMPC local planner. Because the resulting forces are realized
indirectly through kinematic UAV trajectories, an admittance filter further
adjusts the position, velocity, and acceleration references using onboard
cable-tension estimates. Simulation studies with teams of up to ten UAVs and
experiments with four UAVs demonstrate that the proposed approach reduces tension
imbalance and improves robustness to geometric model mismatch while preserving
accurate payload tracking during both hovering and agile flight.
Future work will use the inverse condition number \(k(\mathbf{A})\) to tune the
admittance parameters online, enabling stronger tension correction near
degenerate configurations and less intervention when the allocation matrix is
well conditioned. We will also study how the admittance correction can preserve
the constraints imposed by the local planner.

\addtolength{\textheight}{-12cm}   






\bibliographystyle{IEEEtran}
\bibliography{reference}



\end{document}